\documentclass[runningheads]{llncs}
\usepackage{hyperref}
\usepackage{amssymb}
\usepackage{url}
\usepackage[inline]{enumitem}
\usepackage{varwidth}
\usepackage{capt-of}
\usepackage{wrapfig,lipsum,booktabs}
\usepackage{comment}
\usepackage{xspace}
\usepackage{amsmath}
\usepackage{amssymb}
\usepackage{graphicx}
\usepackage{cite}
\usepackage{multirow}
\usepackage{bbding}
\usepackage{pifont}
\usepackage{xcolor}
\usepackage[normalem]{ulem}
\usepackage{hyperref}
\hypersetup{
  colorlinks,
  citecolor=violet,
  linkcolor=red,
  urlcolor=blue
}

\def\eg{\emph{e.g.,}}

\def\ie{\emph{i.e.,}}

\newcommand\blfootnote[1]{%
  \begingroup
  \renewcommand\thefootnote{}\footnote{#1}%
  \addtocounter{footnote}{-1}%
  \endgroup
}

\begin{document}

\title{Deep Laparoscopic Stereo Matching with Transformers}
\titlerunning{Deep Laparoscopic Stereo Matching with Transformers} 
% \author{ Anonymous submission}
% \institute{Paper ID: 1440 }

\author{\textsuperscript{*}Xuelian Cheng\inst{1} \and
\textsuperscript{*}Yiran Zhong\inst{2,3} \and
Mehrtash Harandi\inst{1,4} \and
Tom Drummond\inst{5} \and
Zhiyong Wang\inst{6} \and
Zongyuan Ge \inst{1,7,8(}\Envelope\inst{)}
}
\authorrunning{X. Cheng et al.}
\institute{Faculty of Engineering, Monash University, Melbourne, Australia 
\and SenseTime Research, Shanghai, China 
\and Shanghai AI Laboratory, Shanghai, China  
\and Data61, CSIRO, Australia
\and University of Melbourne, Melbourne, Australia  
\and The University of Sydney, Sydney, Australia 
\and eResearch Centre, Monash University, Melbourne, Australia
\and Monash-Airdoc Research Centre, Melbourne, Australia
\\
\email{zongyuan.ge@monash.edu}, \url{https://mmai.group}
}

\maketitle 
\begin{abstract}
The self-attention mechanism, successfully employed with the transformer structure is shown promise in many computer vision tasks including image recognition, and object detection. Despite the surge, the use of the transformer for the problem of stereo matching remains relatively unexplored. 
In this paper, we comprehensively investigate the use of the transformer for the problem of stereo matching, especially for laparoscopic videos, and propose a new hybrid deep stereo matching framework (HybridStereoNet) that combines the best of the CNN and the transformer in a unified design. To be specific, we investigate several ways to introduce transformers to volumetric stereo matching pipelines by analyzing the loss landscape of the designs and in-domain/cross-domain accuracy. Our analysis suggests that employing transformers for feature representation learning, while using CNNs for cost aggregation will lead to faster convergence, higher accuracy and better generalization than other options. Our extensive experiments on Sceneflow, SCARED2019 and dVPN datasets demonstrate the superior performance of our HybridStereoNet.

\end{abstract}

\blfootnote{* Indicates equal contribution}

\section{Introduction}
\label{sec:introduction}

3D information and stereo vision are important for robotic-assisted minimally invasive surgeries (MIS)~\cite{cartucho2021visionblender,long2021dssr}. Given the success of modern deep learning systems~\cite{Zhong_2018_ECCV, Zhong_2022_IJCV,Zhong_2018_ECCV_2,cheng2020hierarchical,cheng2019noise} on natural stereo pairs, a promising next challenge is surgical stereo vision, \eg~laparoscopic and endoscopic images. It has received substantial prior interest as its promise for many medical down-streaming tasks such as surgical robot navigation~\cite{overley2017navigation}, 3D registration~\cite{cartucho2021visionblender,long2021dssr}, augmented reality (AR)~\cite{nicolau2011augmented,zhongicpr18} and virtual reality (VR)~\cite{chong2021virtual}.

In recent years, we have witnessed a substantial progress of deep stereo matching in natural images such as %KITTI, Middlebury and ETH3D.
\href{http://www.cvlibs.net/datasets/kitti/eval_scene_flow.php?benchmark=stereo}{KITTI 2015}, \href{http://vision.middlebury.edu/stereo/eval3/}{Middlebury} and \href{https://www.eth3d.net/overview}{ETH3D}. 
Several weaknesses of conventional stereo matching algorithms (\eg~handling occlusion~\cite{scharstein2014high}, and textureless areas~\cite{Menze2015CVPR}) have been largely alleviated through deep convolutional networks~\cite{cheng2020hierarchical,lipson2021raft,NEURIPS2020_add5aebf,Zhong_2019_CVPR,Wang_2021_CVPR} and large training data. However, recovering dense depth maps for laparoscopic stereo videos is still a non-trivial task. First, the textureless problem in laparoscopic stereo images is much severe than natural images. Greater demands were placed on the stereo matching algorithms to handle large textureless areas.
Second, there are only few laparoscopic stereo datasets for training a stereo network due the hardness of retrieving the ground truth. Lack of large-scale training data requires the network to be either an effective learner (\ie~to be able to learn stereo matching from few samples) or a quick adapter that can adjust to the new scene with few samples. Also, there is a  large domain gap between natural images  and medical images. In order to be a quick adapter, the network needs superior generalisation ability to mitigate the gap.

\begin{figure}[t]
    \centering
    \includegraphics[width=1\linewidth]{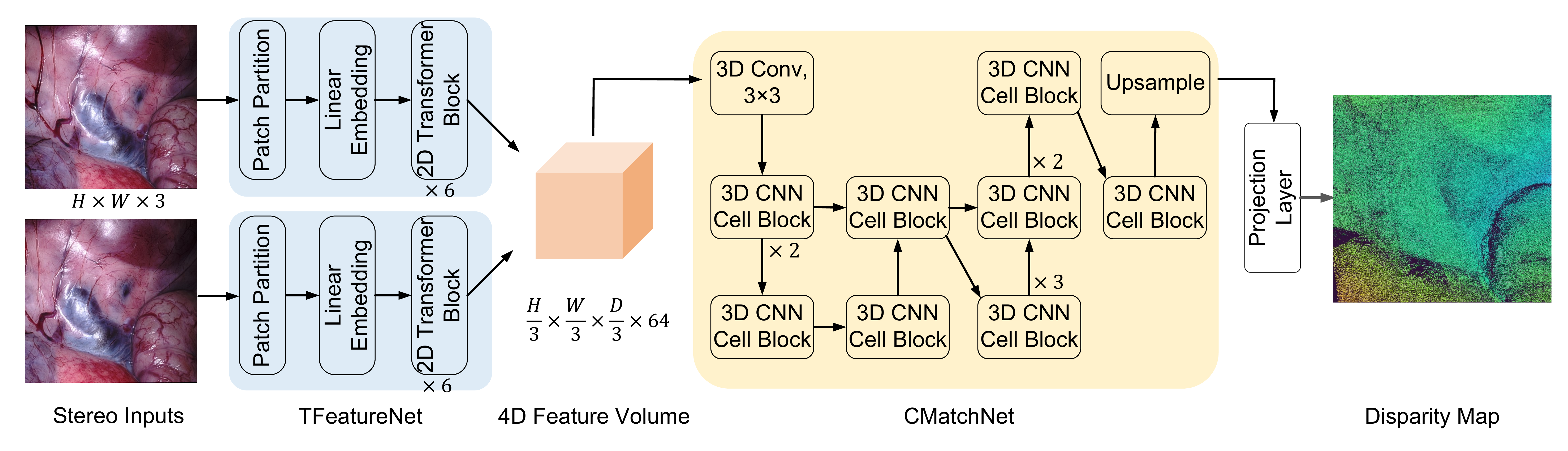}
    \caption{ \small{The overall pipeline of the HybridStereoNet network.} }
    \label{fig:Transtereo_v}
\end{figure}

The transformer has become the horsepower of neural architectures recently, due to its generalization ability~\cite{Sun2022,liu2021swin,wang2021pyramid}. Transformers have been successfully applied in natural language processing~\cite{zhen2022cosformer} and high-level computer vision tasks, \eg~image classification, semantic segmentation and object detection. Yet, its application to low level vision is yet to be proven, \eg~the performance of STTR~\cite{li2020revisiting} is far behind convolution-based methods in stereo matching.
In this paper, we investigate the use of transformers for deep stereo matching in laparoscopic stereo videos. We will show that by using transformers to extract features, while employing convolutions for aggregation of  matching cost,  a deep model with higher domain specific and cross domain performances can be achieved. 

Following the volumetric deep stereo matching pipeline in LEAStereo~\cite{cheng2020hierarchical}, our method consists of a feature net, a 4D cost volume, a matching net and a projection layer. The feature net and the matching net are the only two modules that contain trainable parameters.  We substitute these modules with our designed transformer-based structure and compare the accuracy, generalization ability, and the loss landscapes to analyze the behavior of the transformer for the stereo matching task. 
The insights obtained from those analyses have led us to propose a new hybrid architecture for laparoscopic stereo videos that achieves better performance than both convolution-based methods and pure transformer-based methods. 
Specifically, we find that the transformers tend to find flatter local minima while the CNNs can find a lower one but sharper. By using transformers as the feature extractor, and CNNs for cost aggregation, the network can find a local minima both lower and flatter than pure CNN-based methods, which leads to better generalization ability. Thanks to the global field of view of transformers, our HybridStereoNet has better textureless handling capability as well.

\section{Method}
\label{sec:method}
In this section, we first illustrate our hybrid stereo matching architecture and then provide a detailed analysis of transformers in stereo matching networks. Our design is inspired by LEAStereo which is an state-of-the-art model for the natural stereo matching task. For the sake of discussion, we denote the feature net in LEAStereo \cite{cheng2020hierarchical} as CFeatureNet, and the matching net as CMatchNet.

\begin{figure}[t]
    \centering
    \includegraphics[width=0.95\linewidth]{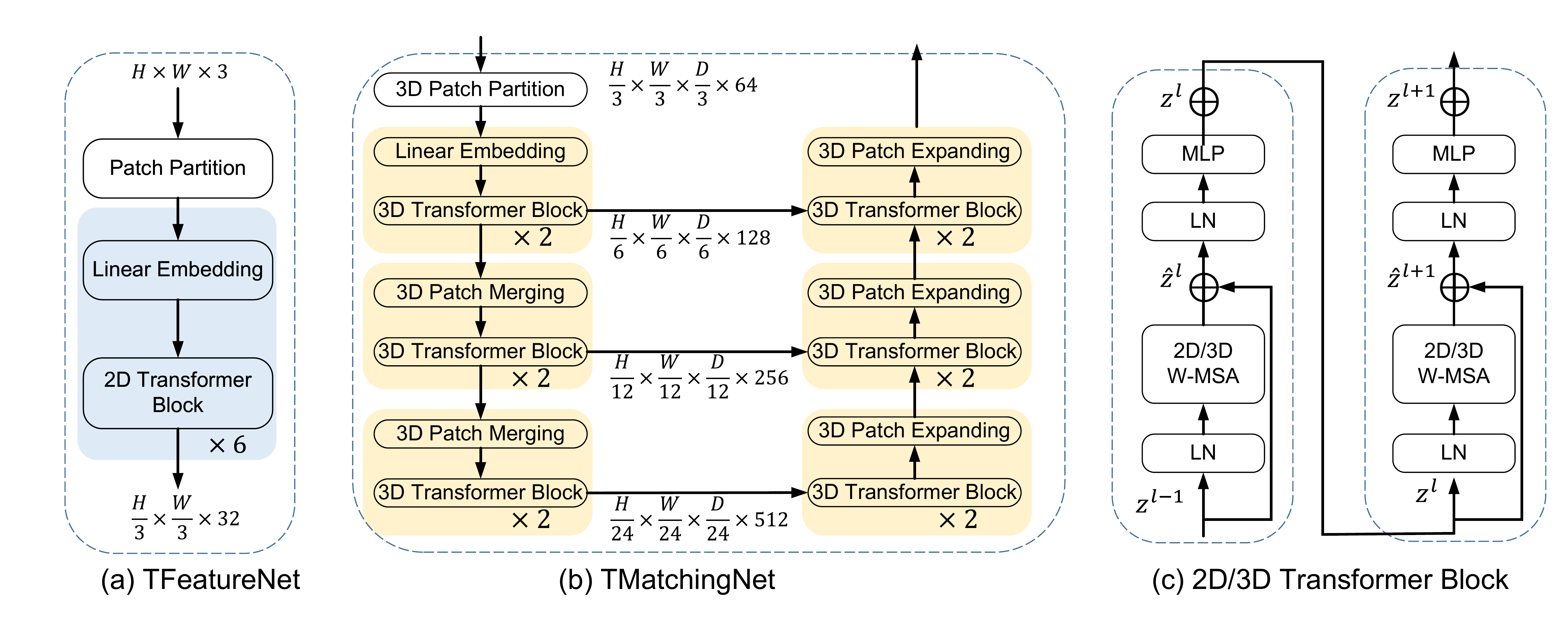}
    \caption{\small{The architecture of our transformer-based TFeatureNet and TMatchNet.}}
    \label{fig:Purestereo}
\end{figure}

\subsection{The HybridStereoNet}
Our proposed HybridStereoNet architecture is shown in Fig~\ref{fig:Transtereo_v}. We adapt the standard volumetric stereo matching pipeline that consists of a feature net to extract features from input stereo images, a 4D feature volume that is constructed by concatenating features from stereo image pairs through epipolar lines~\cite{zhong2017selfsupervised}, a matching net that regularizes the 4D volume to generate a 3D cost volume, and a projection layer to project the 3D cost volume to a 2D disparity map. In this pipeline, only the feature net and the matching net contain trainable parameters. We replace both networks with transformer-based structures to make them suitable for laparoscopic stereo images.

We show our transformer-based feature net (TFeatureNet) in Fig.~\ref{fig:Purestereo} (a) and matching net (TMatchNet) in Fig.~\ref{fig:Purestereo} (b). For a fair comparison with the convolutional structure of the LEAStereo~\cite{cheng2020hierarchical}, we use the same number of layers $L$ for our TFeatureNet and TMatchNet as in the LEAStereo, \ie~$L^F=6$ for the TFeatureNet and $L^M=12$ for the TMatchNet. The 4D feature volume is also built in 1/3 resolution. 

\noindent\textbf{TFeatureNet}
is a Siamese network with shared weights to extract features from input stereo pairs of size $H \times W$. We adapt the same patching technique as in ViT~\cite{dosovitskiy2020image} and split the image into $N$ non-overlapping patches before feeding them to a vision transformer. We set the patch size to $3 \times 3$ to obtain $\frac{H}{3} \times \frac{W}{3}$ tokens and thus ensure the cost volume built in 1/3 resolution. A linear embedding layer is applied on the raw-valued features and project them to a $C$ dimensional space. We set the embedding feature channel to $32$. We empirically select the swin transformer block~\cite{liu2021swin} as our 2D transformer block.

\noindent\textbf{TMatchNet} is a U-shaped encoder-decoder network with 3D transformers. The encoder is a three-stage down-sampling architecture with stride of 2. In contrast, the decoder is a three-stage up-sampling architecture as shown in Fig~\ref{fig:Purestereo} (b). Similar to the TFeatureNet, we use a 3D Patch Partitioning layer to split the 4D volume $H' \times W' \times D' \times C$ into $N$ non-overlapping 3D patches and feed them into a 3D transformer. Unlike previous 3D transformers~\cite{liu2021video} that keep the dimension $D$ unchanged when down-sampling the spatial dimensions $H,W$, we change $D$ accordingly with the stride to enforce the correct geometrical constraints.
\begin{figure}[t]
    \centering
    \includegraphics[width=1\linewidth]{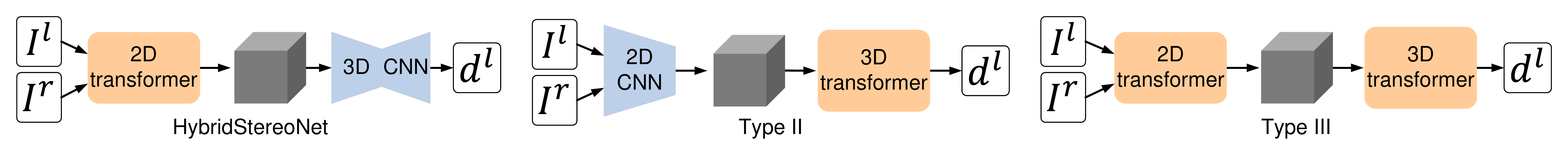}
    \caption{\small{The overall pipeline of variant networks. We follow the feature extraction – 4D feature volume construction - dense matching pipeline for deep stereo matching. The variants change the FeatureNet and Matching Net with either transformer or CNNs. The gray box represents 4D cost volume.}}
    \label{fig:Variants}
\end{figure}

We compare the functionality of transformers in feature extraction and cost aggregation. As we will show soon, we find that transformers are good for feature representation learning while convolutions are good for cost aggregation. Therefore, in our HybridStereoNet, we use our TFeatureNet as the feature extractor and CMatchNet for the cost aggregation. We provide a detailed comparsion and analysis in the following section.

\subsection{Analyzing transformer in laparoscopic stereo}
To integrate the transformer in the volumetric stereo pipeline, we have three options as shown in Fig~\ref{fig:Variants}. Type I (HybridStereoNet): TFeatureNet with CMatchNet; Type II: CFeatureNet with TMatchNet; and Type III: TFeatureNet with TMatchNet.
In this section, we investigate these options in terms of loss landscapes, projected learning trajectories, in-domain/cross-domain accuracy, and texture-less handling capabilities in laparoscopic stereo.

\noindent\textbf{Loss landscape.}
We visualize the loss landscape on the SCARED2019 dataset for types described above along with LEAStereo using the random direction approach~\cite{li2018visualizing} in Fig~\ref{fig:loss_lanscape}. The plotting details can be found in the supplemental material. The visualization suggests that Type II and Type III tend to find a flatter loss landscape but with a higher local minima. Compared with LEAStereo, the landscape of the HybridStereoNet shares a similar local minima but is flatter,  which leads to better cross-domain performance~\cite{chaudhari2019entropy, keskar2016large}.

\begin{figure*}[t!]
\small
    \centering
    \tabcolsep=0.02cm
    \renewcommand{\arraystretch}{1.0}
    \begin{tabular}{ c c c c }
    \includegraphics[width=0.225\linewidth]{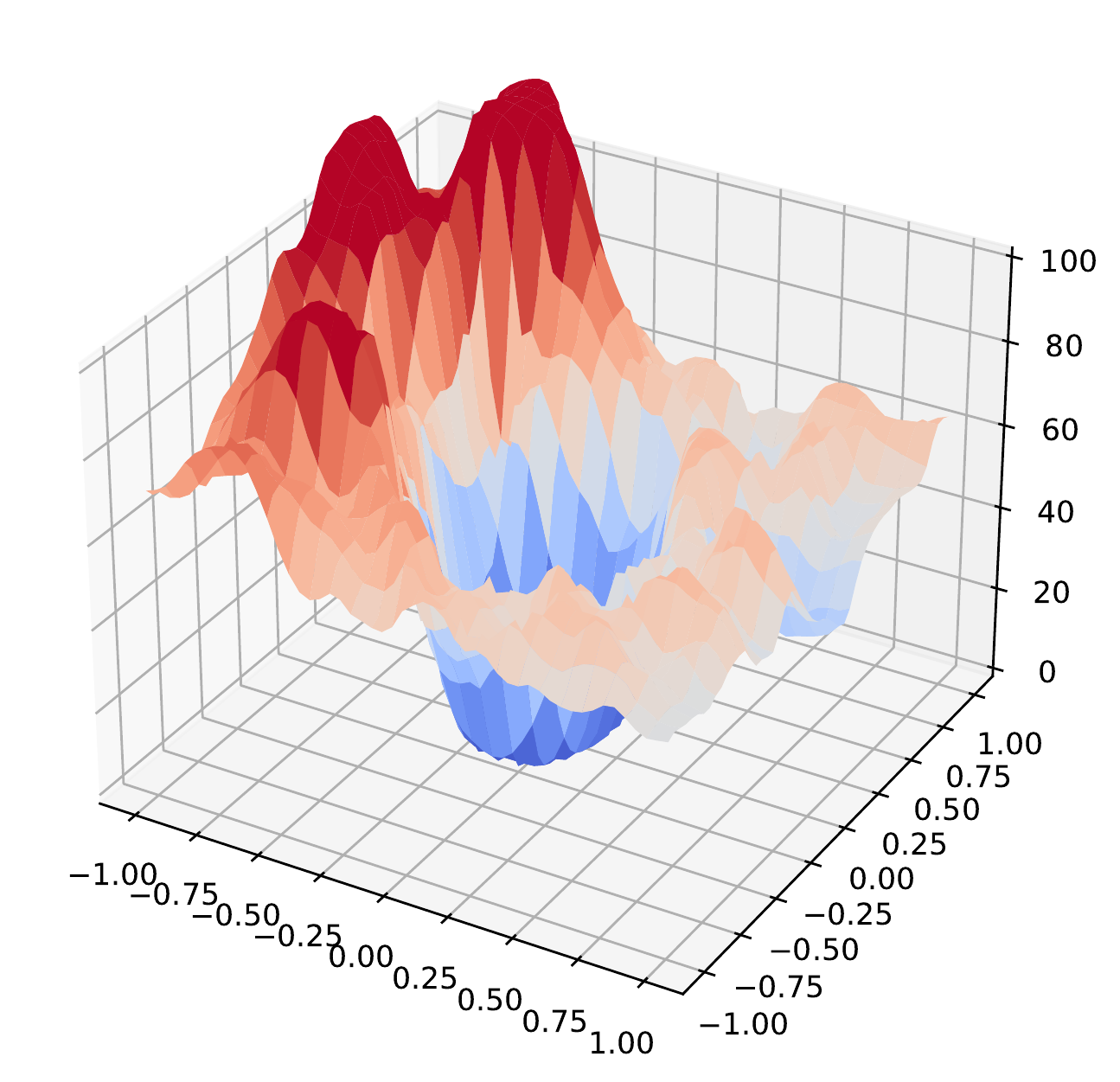} &
    \includegraphics[width=0.225\linewidth]{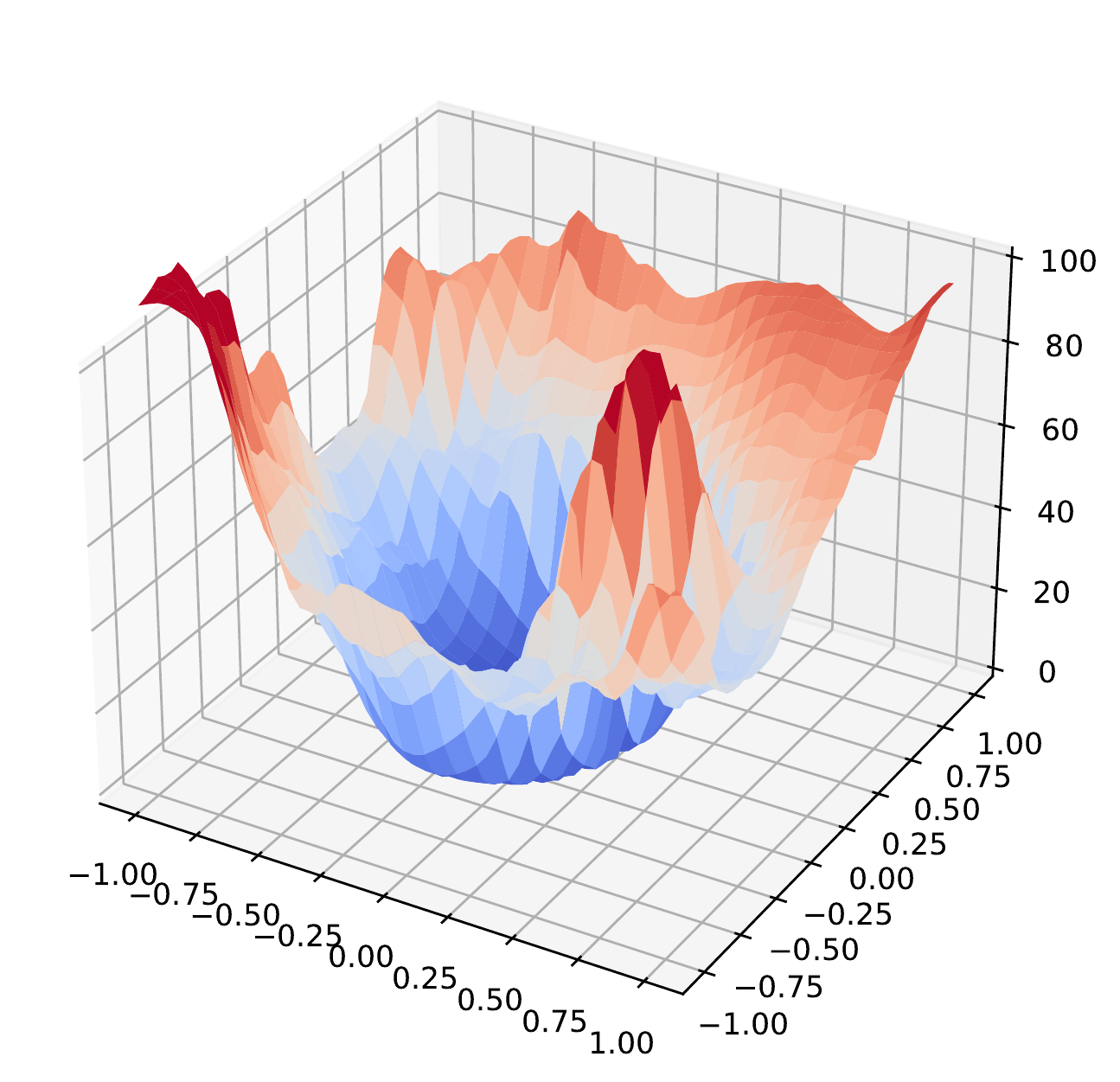} &
    \includegraphics[width=0.225\linewidth]{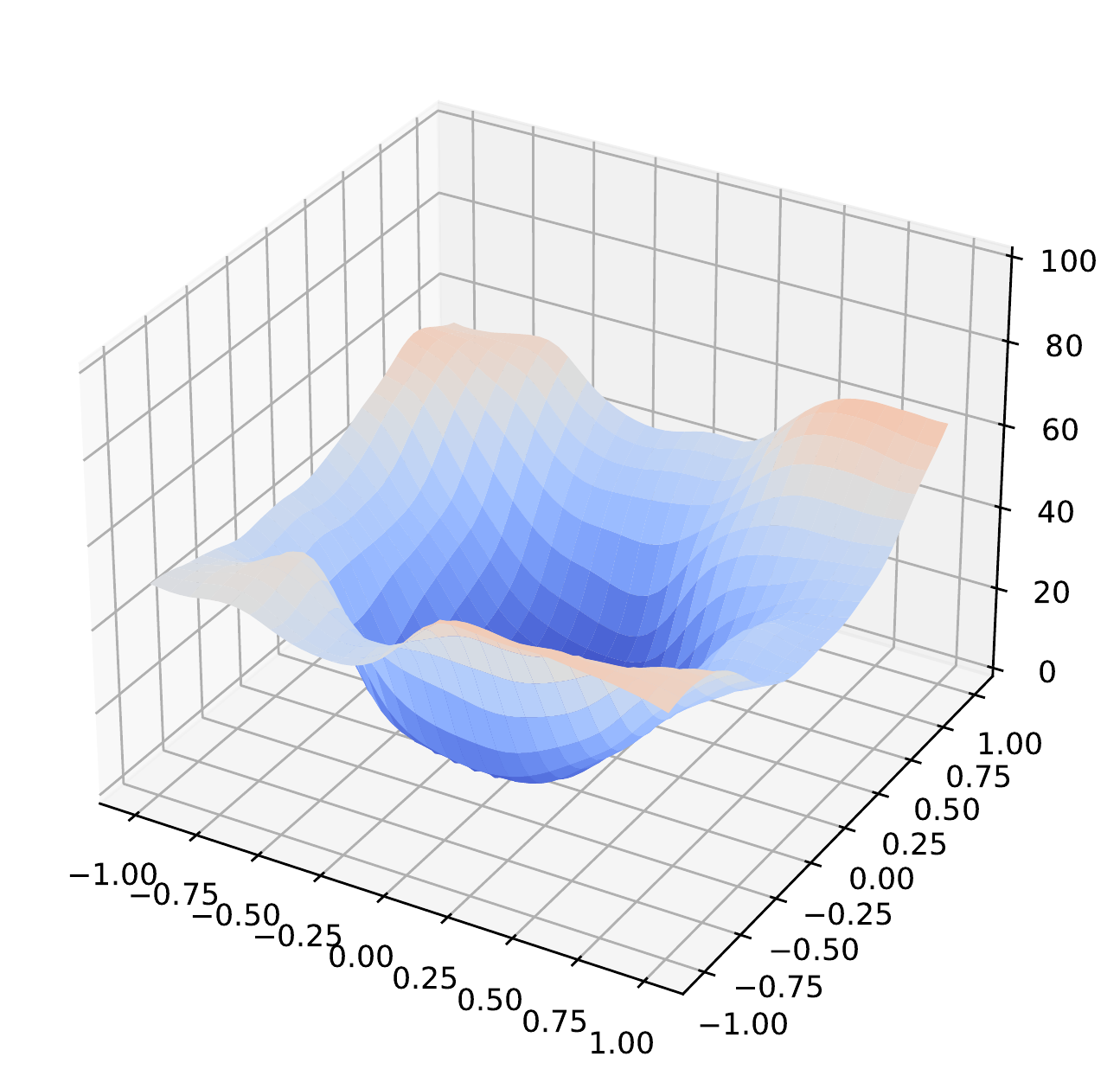} &
    \includegraphics[width=0.259\linewidth]{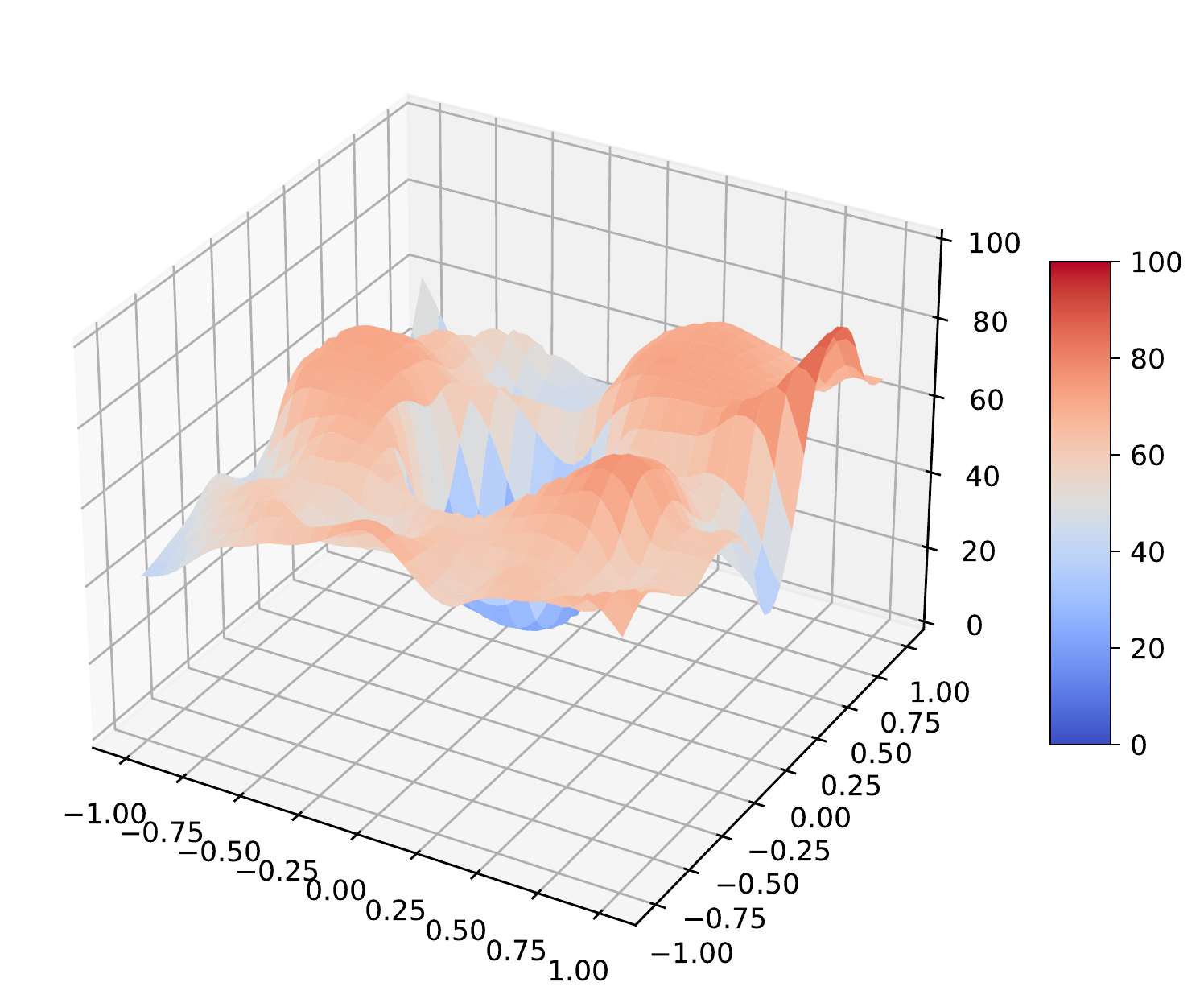} \\
    \small{LeaStereo} & \small{HybridStereoNet} & \small{Type II} & \small{Type III} \\
    \end{tabular}
    \caption{\small{Loss Landscape Visualization on SCARED2019 dataset. 3D surfaces of the gradient variance from HybridStereoNet and its variants on SceneFlow. The two axes mean two random directions with filter-wise normalization. The height of the surface indicates the value of the gradient variance.}}
    \label{fig:loss_lanscape}
\end{figure*}

\begin{figure*}[b]
\centering
\begin{tabular}{ c c c c }
\hspace{-.5cm}
 \includegraphics[width=0.245\linewidth]{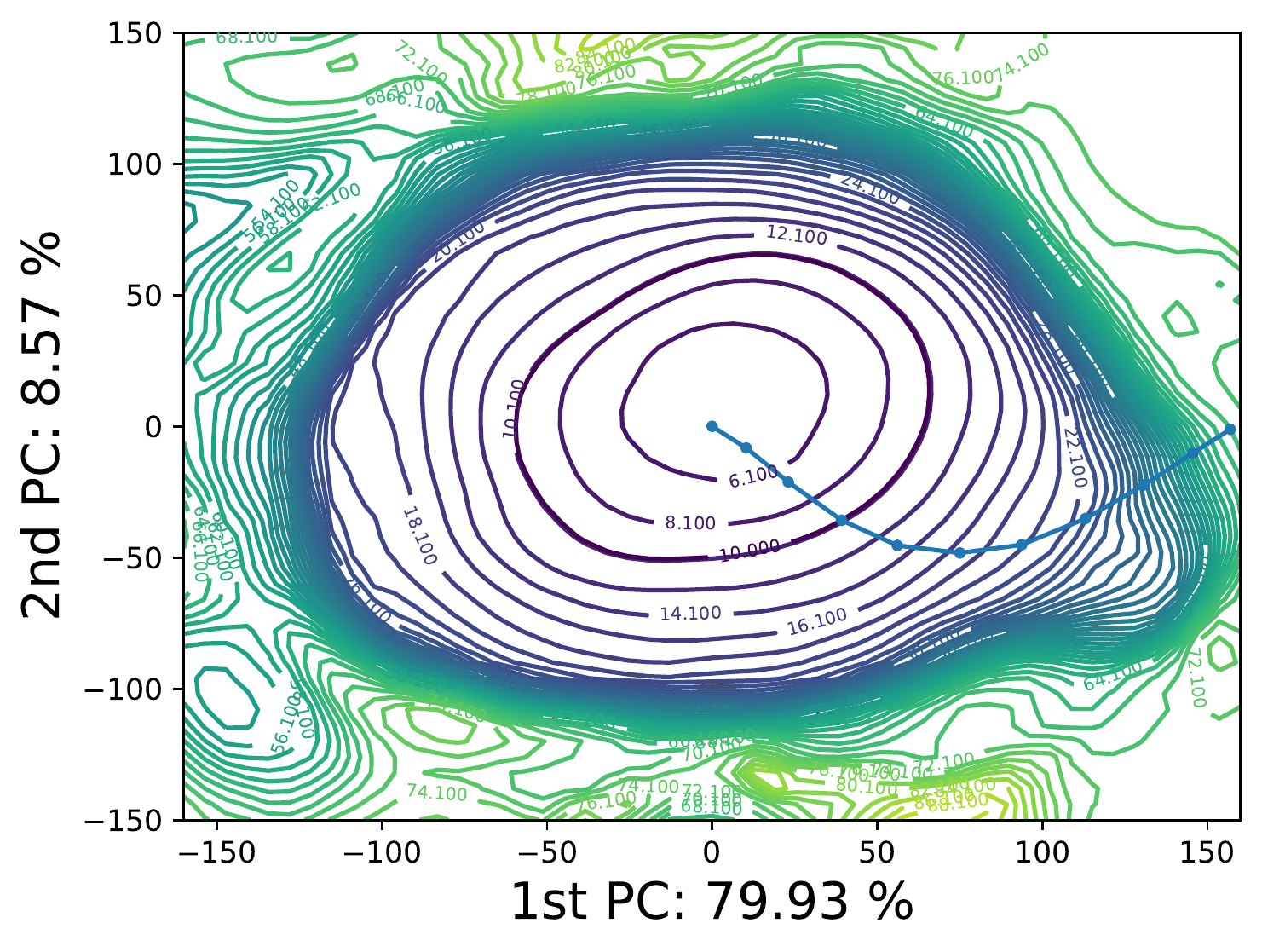} & 
 \includegraphics[width=0.245\linewidth]{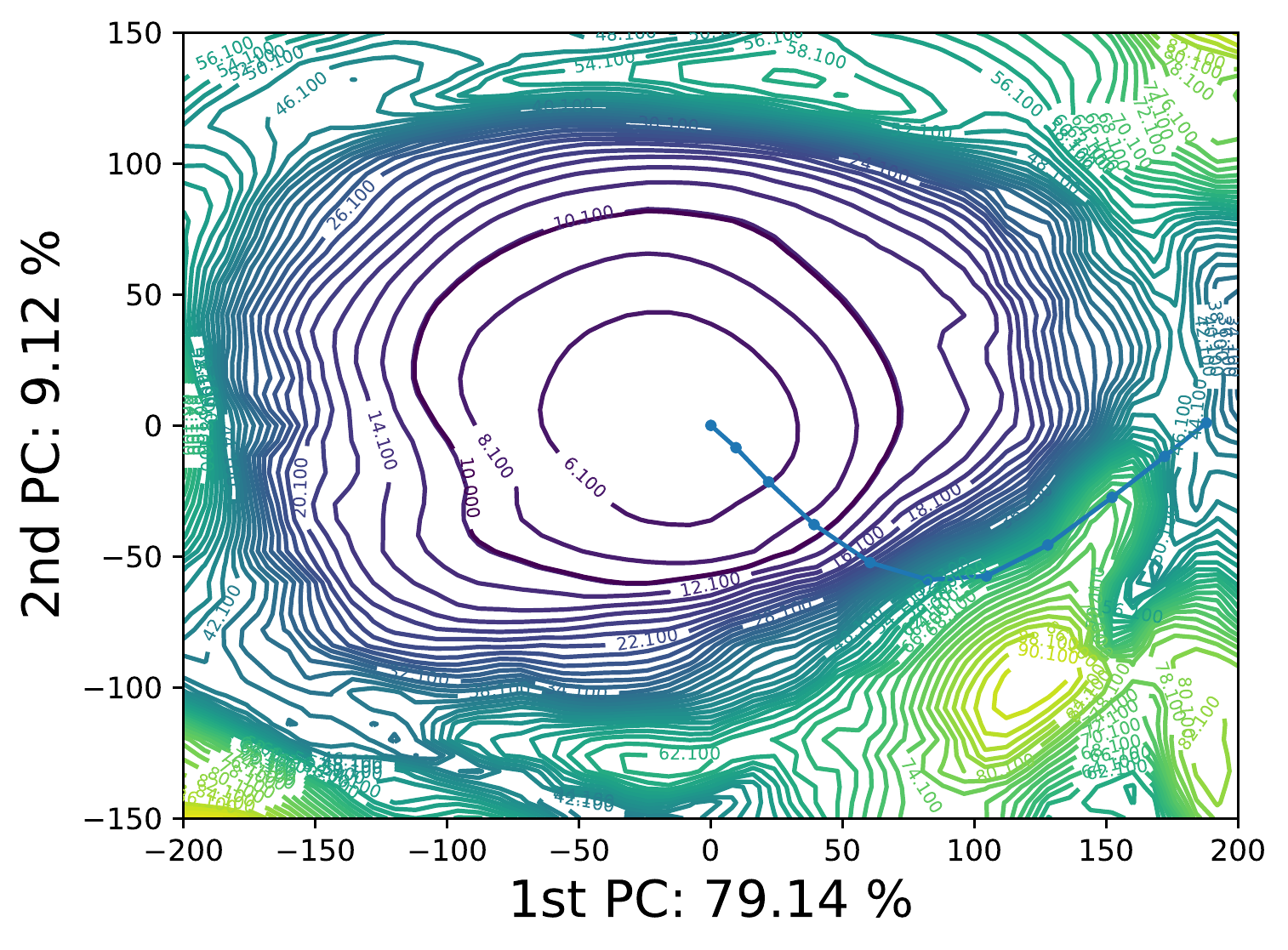}  & 
 \includegraphics[width=0.245\linewidth]{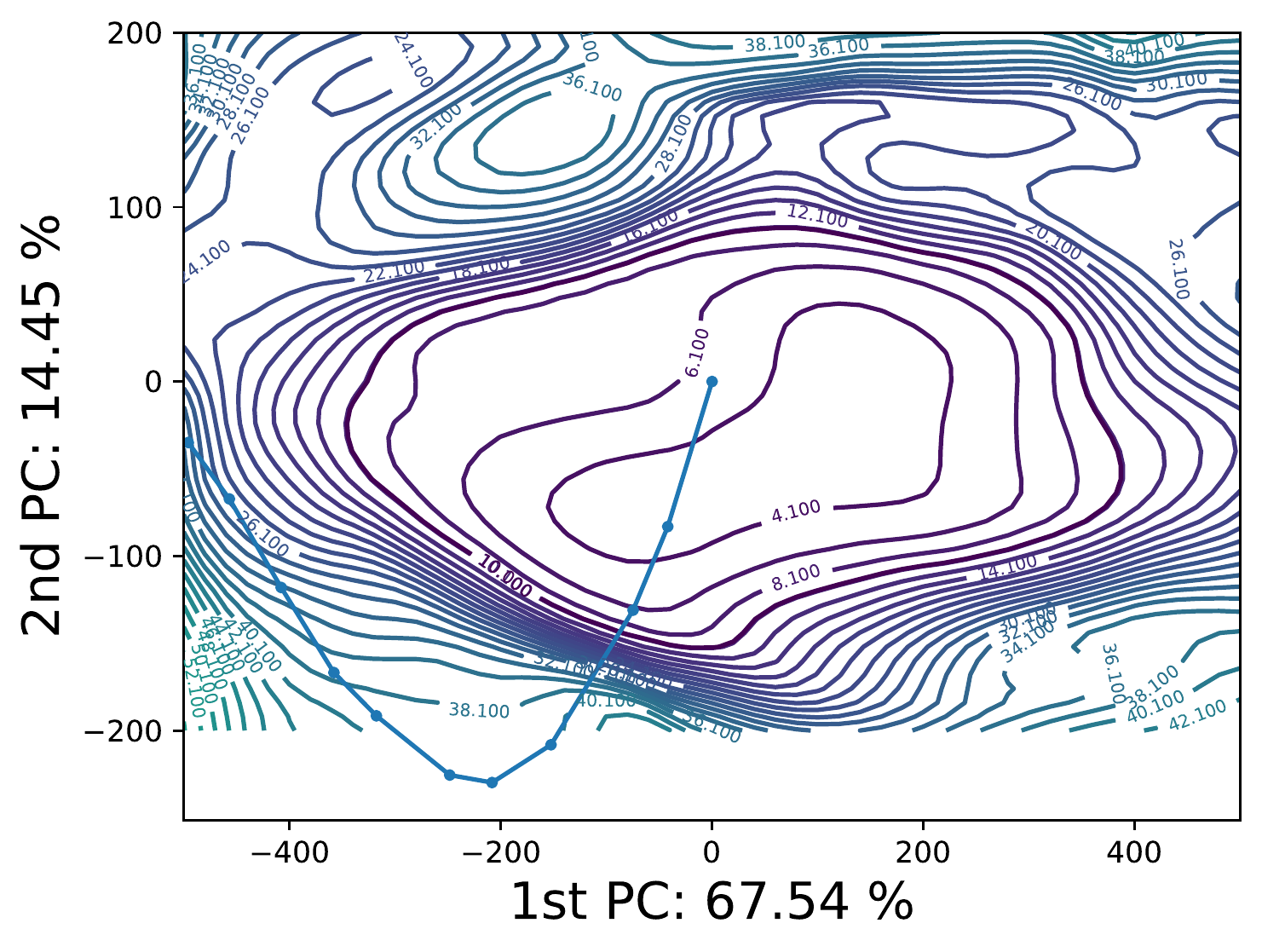} & 
 \includegraphics[width=0.245\linewidth]{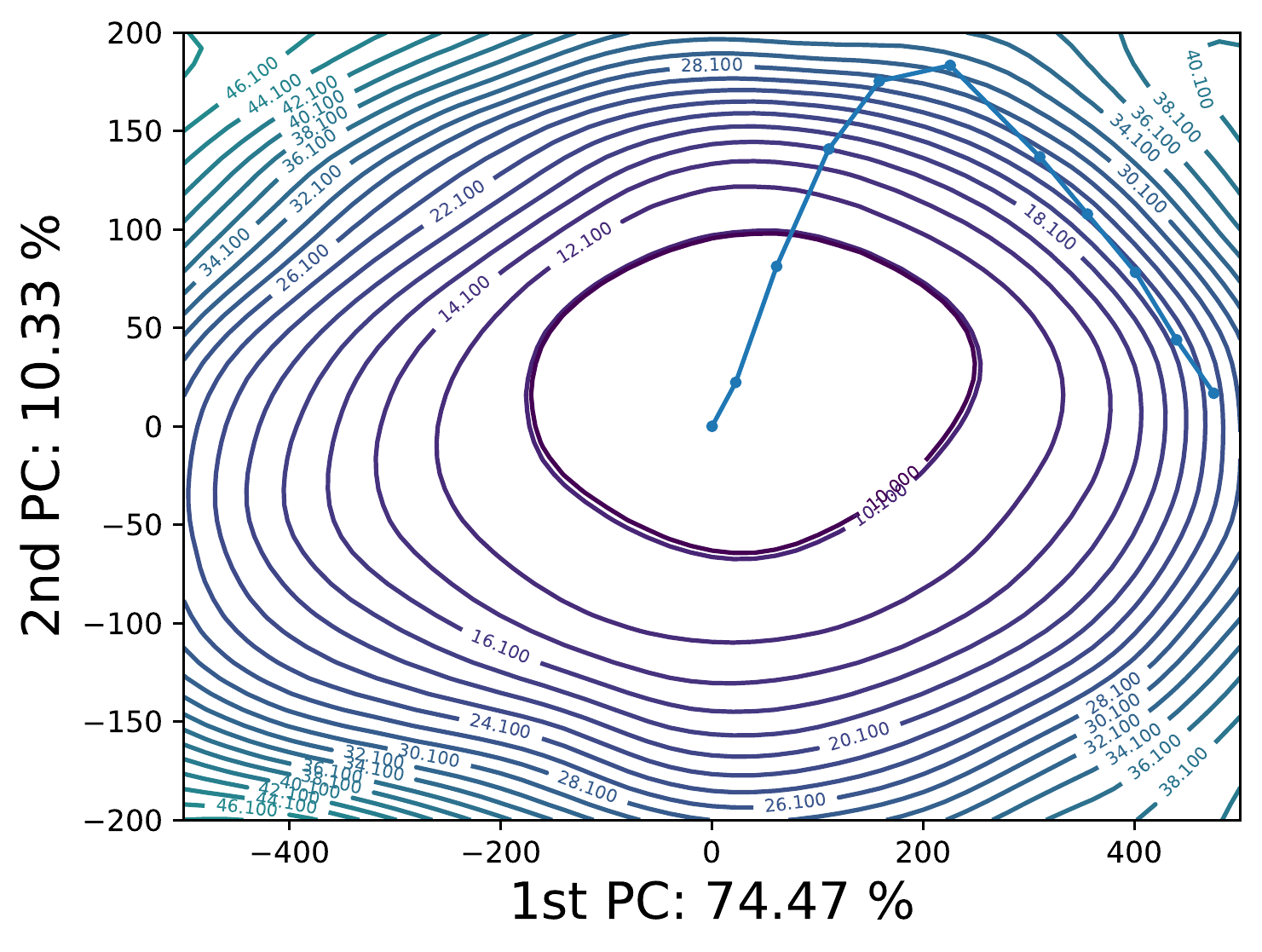}
 \\
 LEAStereo & HybridStereoNet & Type II & Type III \\
\end{tabular}
\caption{\small{Projected learning trajectories use normalized PCA directions for our variants. Please zoom in for better visualization.}}
\label{fig:pca_projections}
\end{figure*}

\noindent\textbf{Projected learning trajectory.}
We also compare the convergence curve of each type with the projected learning trajectory~\cite{li2018visualizing}. Fig.~\ref{fig:pca_projections} shows the learning trajectories along the contours of loss surfaces. Let $\theta_i$ denote model parameters at epoch $i$. The final parameters of the model after $n$ epochs of training are shown by $\theta_n$. Given $n$ training epochs, we can apply PCA~\cite{scholkopf1998nonlinear} to the matrix
$M = [\theta_0 - \theta_n; \cdots; \theta_{n-1} - \theta_n],$
and then select the two most explanatory directions.
This enables us to visualize the optimizer trajectory (blue dots in Fig.~\ref{fig:pca_projections}) and loss surfaces along PCA directions. On each axis, we measure the amount of variation in the descent path captured by that PCA direction. Note that the loss landscape dynamically changes during training and we only present the ``local'' landscape around the final solution.

As we can see from the Fig.~\ref{fig:pca_projections}, the Type II variant misses the local minima; Type III directly heads to a local minima after circling around the loss landscape in its first few epochs. It prohibits the network from finding a lower local minima. The LEAStereo and our HybridStereoNet both find some lower local minimas but the HybridStereoNet uses a sharper descending pathway on the loss landscape and therefore leads to a lower local minima. We use the SGD optimizer and the same training setting with LEAStereo \cite{cheng2020hierarchical} for side-by-side comparison. %Noting that a different optimizer could lead to a different minimum with different behavior. We may extend a more thorough investigation, \ie~randomness in the loss landscape, as our future work.

\begin{figure}[t]
    \centering
    \includegraphics[width=0.45\linewidth]{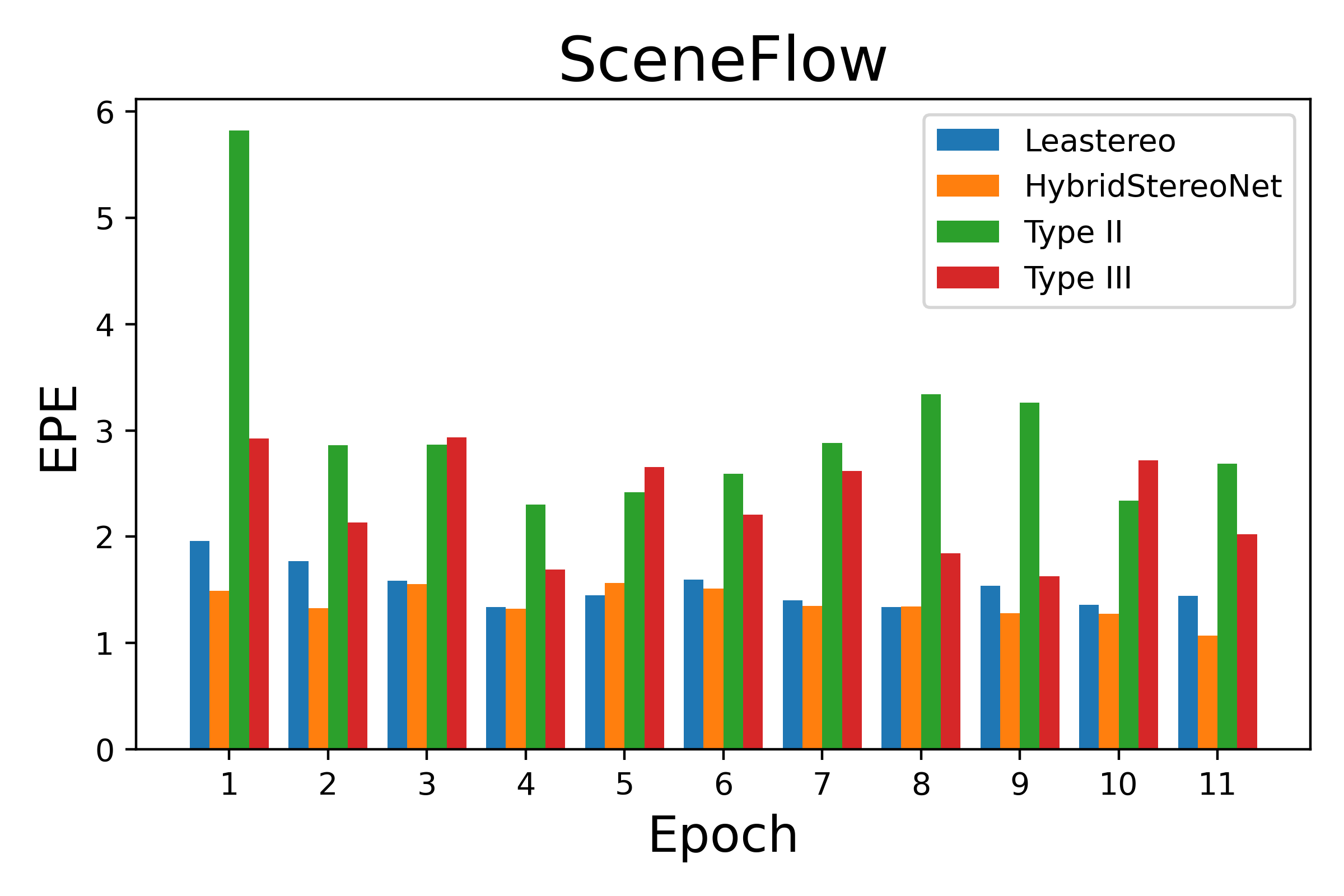} 
    \includegraphics[width=0.45\linewidth]{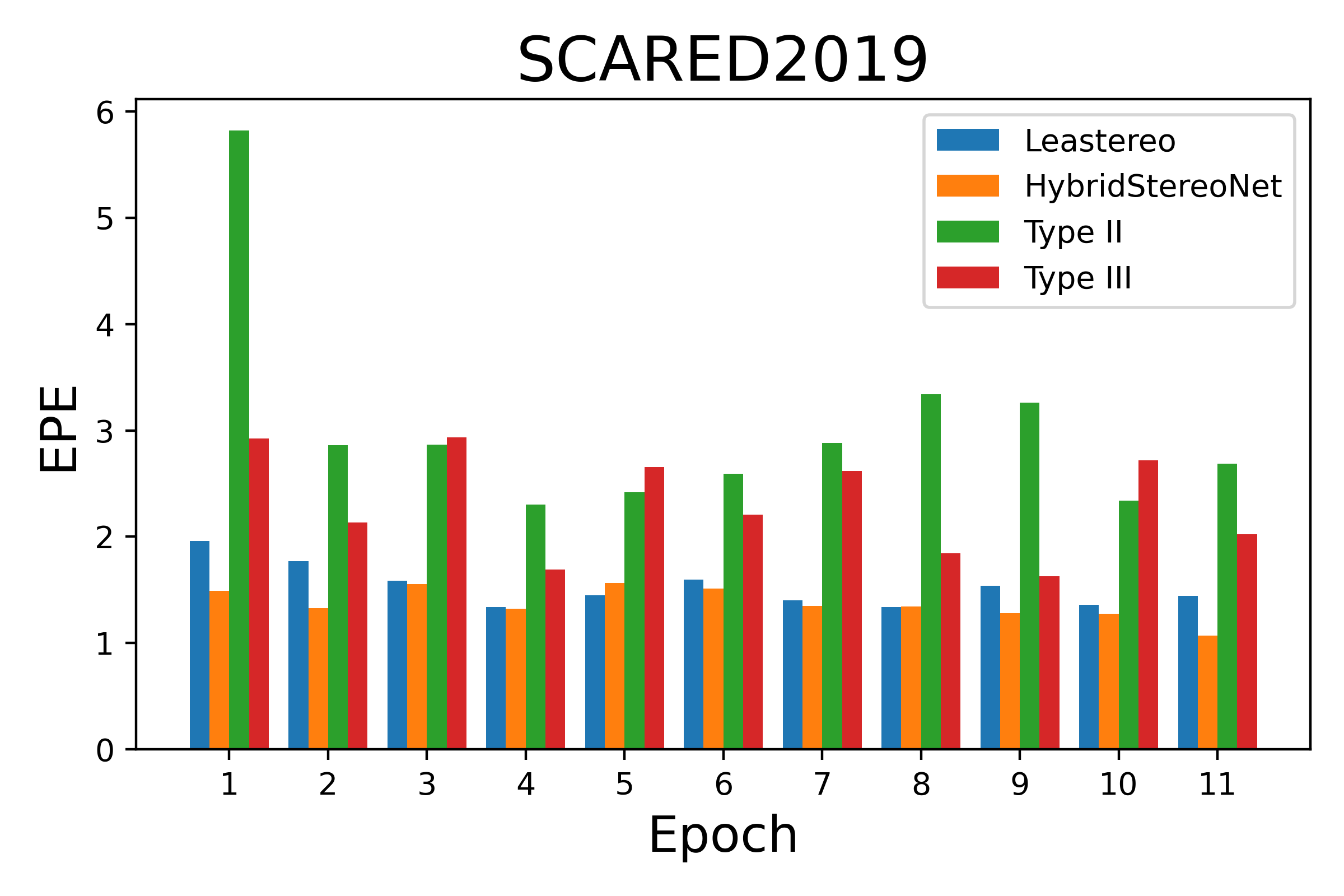}
    \caption{\small{\textbf{ Accuracy comparison for our proposed variants.} (a) In-domain results on SceneFlow dataset. (b) Cross domain results on SCARED2019 dataset.} }
    \label{fig:acc}
\end{figure}

\noindent\textbf{Accuracy.}
In Fig~\ref{fig:acc}, we compare the in-domain and cross-domain  accuracy of all types for the first several epochs. All of these variants are trained on a large nature scene synthetic stereo dataset called SceneFlow~\cite{Mayer2016CVPR} with over 30k stereo pairs. For in-domain performance, we plot the validation end point error (EPE) on the SceneFlow dataset in Fig~\ref{fig:acc} (a). The HybridStereoNet consistently achieves low error rates than all the other variants. For the cross-domain performance, we directly test the trained models on the SCARED2019 laparoscopic stereo dataset and plot the EPE in Fig~\ref{fig:acc} (b). Again, the HybridStereoNet achieves lower cross-domain error rates in most epochs.

\begin{table*}[t]
\scriptsize
\centering
\tabcolsep=0.25cm
\caption{Memory footprint for the proposed variant networks.} 
\begin{tabular}{l c c c c }
\toprule
Method  & LeaStereo  & Type II & Type III & HybridStereoNet \\
\hline
Params [M]  & 1.81 & 9.54 & 9.62 & 1.89 \\ 
Flops [M]   & 3713.53 & 4144.85 & 811.68 & 380.01 \\
Runtime [s] & 0.30 & 0.48 & 0.50 & 0.32 \\
\bottomrule
\end{tabular}
\label{tab:memory} 
\end{table*}

\noindent\textbf{Memory footprint.} We provide details of four variants regarding the running time, and memory footprint in Table \ref{tab:memory}, which were tested on a Quadro GV100 with the input size $504 \times 840$. We keep relatively similar number of learnable parameters to make a fair comparison. %As a part of the work, we also investigated different options of learnable parameters for the proposed variants. The performance is not improved significantly with the parameters increased. For example, pure transformer structures (9.62MB and 28.94MB) share very similar training trends and converge to similar validation accuracy. 

\section{Experiments}
\label{sec:exp}

%% SCARED dataset %%%%%
\begin{table*}[t]
\scriptsize
\centering
\caption{The mean absolute depth error for the SCARED2019 \textit{Test-Original} set (unit: mm). Each test set containing 5 keyframes, denoted as $kf_n, n\in [1,5]$. Note that our method and STTR are not fine-tuned on the target dataset. The lower the better.}
\label{tab:scared_depth}
\begin{tabular}{p{3mm} l cccccc cccccc}
\toprule
& & \multicolumn{6}{c}{Test Set 1} & \multicolumn{6}{c}{Test Set 2} \\
\cmidrule(lr){3-8}
\cmidrule(lr){9-14}
 & Method & $kf_1$ & $kf_2$ & $kf_3$ & $kf_4$ & $kf_5$ & Avg. & $kf_1$ & $kf_2$ & $kf_3$ & $kf_4$ & $kf_5$ & Avg. \\
\midrule
\multirow{11}{*}{\rotatebox{90}{Supervised}} 
& Lalith Sharan~\cite{allan2021stereo} & 30.63 & 46.51 & 45.79 & 38.99 & 53.23 & 43.03 & 35.46 & 50.09 & 25.24 & 62.37 & 70.45 & 48.72 \\
& Xiaohong Li~\cite{allan2021stereo} & 34.42 & 20.66 & 17.84 & 27.92 & 13.00 & 22.77 & 24.58 & 16.80 & 29.92 & 11.37 & 19.93 & 20.52 \\
& Huoling Luo~\cite{allan2021stereo} & 29.68 & 16.36 & 13.71 & 22.42 & 15.43 & 19.52 & 20.83 & 11.27 & 35.74 & 8.26 & 14.97 & 18.21 \\
& Zhu Zhanshi~\cite{allan2021stereo} & 14.64 & 7.77 & 7.03 & 7.36 & 11.22 & 9.60  & 14.41 & 12.55 & 16.30 & 27.87 & 34.86 & 21.20 \\
& Wenyao Xia~\cite{allan2021stereo} & \textbf{5.70} & 7.18 & 6.98 & 8.66 & 5.13 & 6.73 & 13.80 & 6.85 & 13.10 & 5.70 & 7.73 & 9.44 \\
& Trevor Zeffiro~\cite{allan2021stereo} & 7.91 & 2.97 & \textbf{1.71} & 2.52 & 2.91 & 3.60  & 5.39 & 1.67 & 4.34 & 3.18 & 2.79 & 3.47 \\
& Congcong Wang~\cite{allan2021stereo} & 6.30 & {2.15} & 3.41 & 3.86 & 4.80 & 4.10 & 6.57 & 2.56 & 6.72 & 4.34 & 1.19 & 4.28 \\
& J.C. Rosenthal~\cite{allan2021stereo} & 8.25 & 3.36 & 2.21 & {2.03} & 1.33 & 3.44 & 8.26 & 2.29 & 7.04 & 2.22 & 0.42 & 4.05 \\
& D.P. 1~\cite{allan2021stereo} & 7.73 & 2.07 & 1.94 & 2.63 & \textbf{0.62} & 3.00 & 4.85 & \textbf{0.65}& \textbf{1.62} & \textbf{0.77} & 0.41 & \textbf{1.67} \\
&D.P. 2~\cite{allan2021stereo}  & 7.41 & \textbf{2.03} & 1.92 & 2.75 & 0.65 & 2.95 & 4.78 & 1.19 & 3.34 & 1.82 & \textbf{0.36} & 2.30 \\
\hline
% & STTR \cite{li2020revisiting} & 7.24 & 3.93 & 2.49 & \textbf{1.92} & 3.05 & 3.77 & 6.7 & 5.51 & 2.72 & 9.42 & 6.72 & 6.21 \\
% & HybridStereoNet &  6.74 & 3.02 & 2.49 & 3.02 & 1.21 & \textbf{2.95} & 5.58 & 2.4 & 2.56 & 4.47 & 0.53 & 3.13 \\
& STTR \cite{li2020revisiting} & 9.24 & 4.42 & 2.67 & \textbf{2.03} & 2.36 & 4.14 & 7.42 & 7.40 & 3.95 & 7.83 & 2.93 & 5.91  \\
& HybridStereoNet & 7.96 & 2.31 & 2.23 & 3.03 & 1.01 & 3.31 & \textbf{4.57} & 1.39 & 3.06 & 2.21 & 0.52 & 2.35 \\
\hline
\end{tabular}
\end{table*}

\subsection{Datasets}
We evaluate our HybridStereoNet on two public laparoscopic stereo datasets: the SCARED2019 dataset~\cite{allan2021stereo} and the dVPN dataset \cite{ye2017self}. 

\noindent \textbf{SCARED2019} 
is released during the Endovis challenge at MICCAI 2019, including 7 training subsets and 2 test subsets captured by a da Vinci Xi surgical robot. The original dataset only provides the raw video data, the depth data of each key frame and corresponding camera intrinsic parameters. We perform additional dataset curation to make it suitable for stereo matching. After curation, the SCARED2019 contains 17206 stereo pairs for training and 5907 pairs for testing. We use the official code to assess the mean absolute depth error on all the subsequent frames provided by~\cite{allan2021stereo}, named \textit{Test-Original}.

However, as pointed by STTR \cite{li2020revisiting}, the depth of following frames are interpolated by forwarding kinematics information of the point cloud. This would lead to the synchronization issues and kinematics offsets, resulting in inaccurate depth values for subsequent frames. Following STTR \cite{li2020revisiting}, we further collect the first frame of each video and build our \textit{Test-19} set, which subset consists of 19 images of resolution 1080 × 1024 with the maximum disparity of 263 pixels. The left and right 100 pixels were cropped due to invalidity after rectification. We further provide a complete \textit{one-key evaluation toolbox} for disparity evaluation. 

\noindent \textbf{dVPN}
is provided by Hamlyn Centre Laparoscopic, with 34320 pairs of rectified stereo images for training and 14382 pairs for testing.
There is no ground truth depth for these frames. To compare the performance of our model with other methods, we use the image warping accuracy~\cite{zhong2017selfsupervised} as our evaluation metrics, \ie~Structural Similarity Index Measure (SSIM) \cite{wang2004image}, and Peak-Signal-to-Noise Ratio (PSNR) \cite{hore2010image}. Note that for a fair comparison, we exclude self-supervised methods in our comparison as they directly optimize the disparity with image warping losses~\cite{zhong2017selfsupervised}.

\subsection{Implementation}
We implemented all the architectures in Pytorch. A random crop with size $336 \times 336$ is the only data argumentation technique used in this work. We use the SGD optimizer with momentum 0.9, cosine learning rate that decays from 0.025 to 0.001, and weight decay 0.0003. Our pretrained models on SceneFlow are conduced on two Quadro GV100 GPUs. Due to the limitation of public laparoscopic data and the ground truth, we train the proposed variant models on a synthetic dataset, SceneFlow \cite{Mayer2016CVPR}, which has per-pixel ground truth disparities. It contains 35,454 training and 4,370 testing rectified image pairs with a typical resolution of $540 \times 960$. We use the ``finalpass'' version as it is more realistic.

\begin{table*}[t]
\scriptsize
\centering
\tabcolsep=0.18cm
\caption{\small{Quantitative results on the \textit{Test-19} set (evaluated on all pixels). We compare our method with various state-of-the-art methods, by bad pixel ratio disparity errors. }} 
\begin{tabular}{l c c c c c c c c}
\toprule
Methods & EPE [px] $\downarrow$ & RMSE [px] $\downarrow$ & bad 2.0 [\%] $\downarrow$ & bad 3.0 [\%] $\downarrow$ & bad 5.0 [\%] $\downarrow$ \\ 
\hline
STTR \cite{li2020revisiting} & 6.1869 & 20.4903 & 8.4266 &  8.0428 & 7.5234 \\ 
LEAStereo \cite{cheng2020hierarchical} & 1.5224 & \textbf{4.1135} & 4.5251 & 3.6580 & 2.1338 \\
HybridStereoNet & \textbf{1.4096} & 4.1336 & \textbf{ 4.1859} & \textbf{3.4061} & \textbf{2.0125} \\
\bottomrule
\end{tabular}
\label{tab:scared_small} 
\end{table*}

\begin{figure}[t]
\small
    \centering
    \tabcolsep=0.02cm
    \renewcommand{\arraystretch}{1.0}
    \begin{tabular}{c c c c c c}
    %\rotatebox{90}{~SCARED2019} &
    \includegraphics[width=0.195\linewidth]{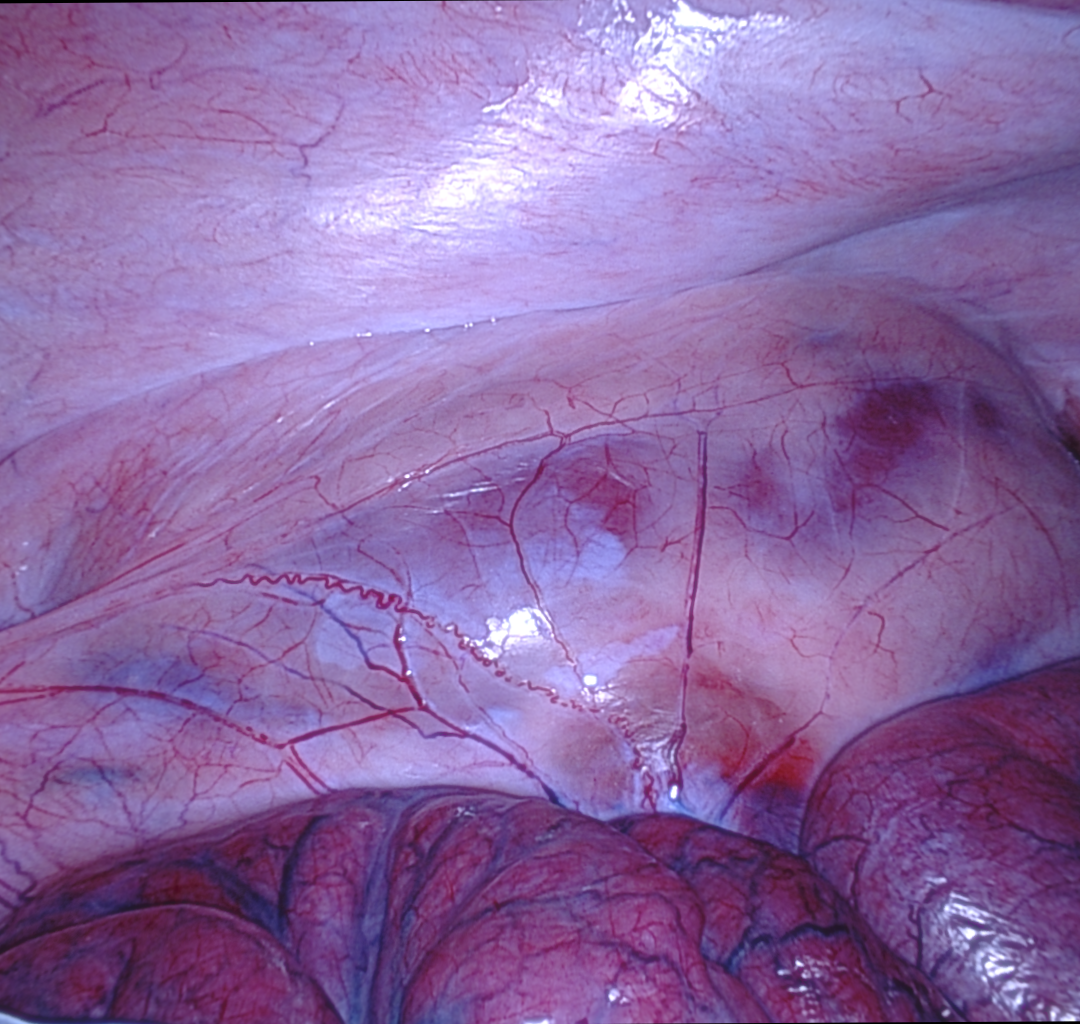} &
    \includegraphics[width=0.195\linewidth]{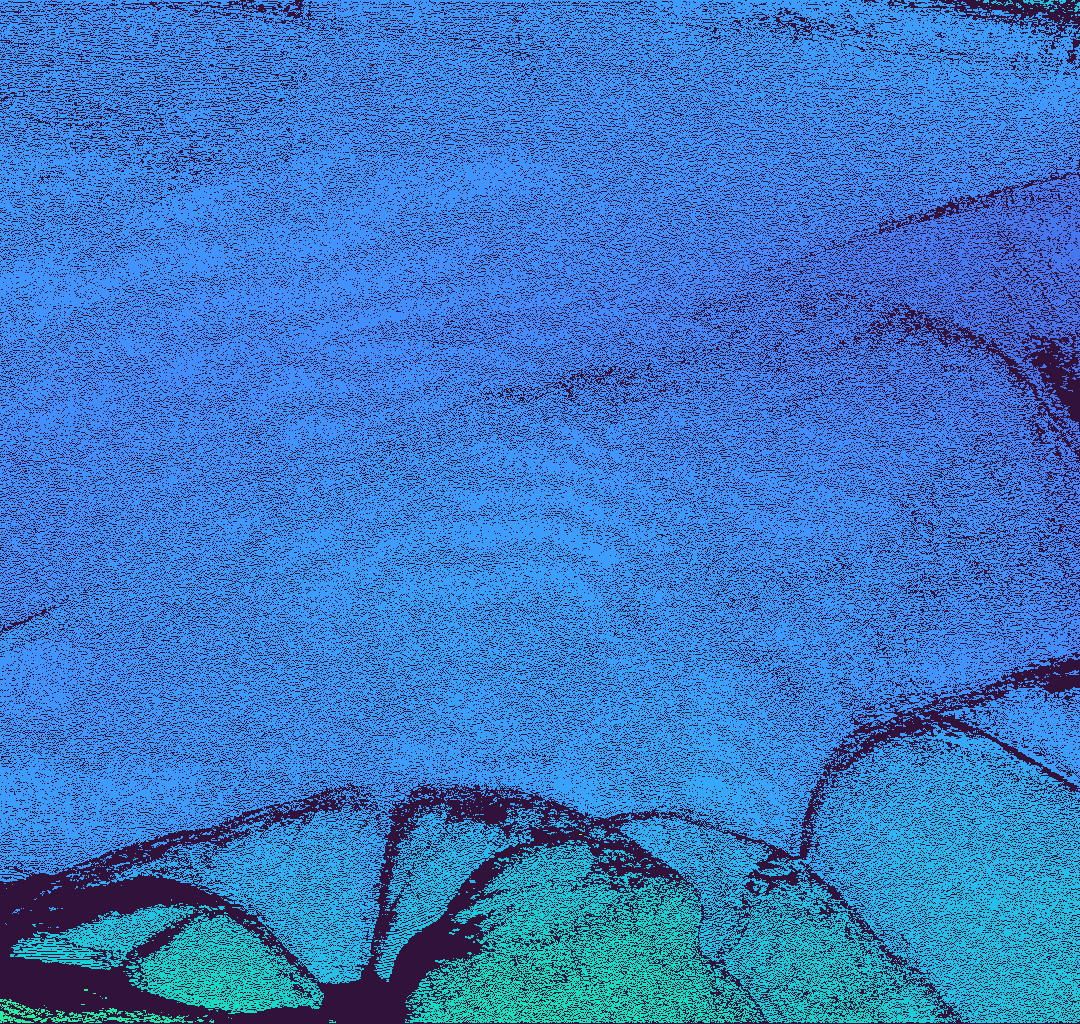} & 
    \includegraphics[width=0.195\linewidth]{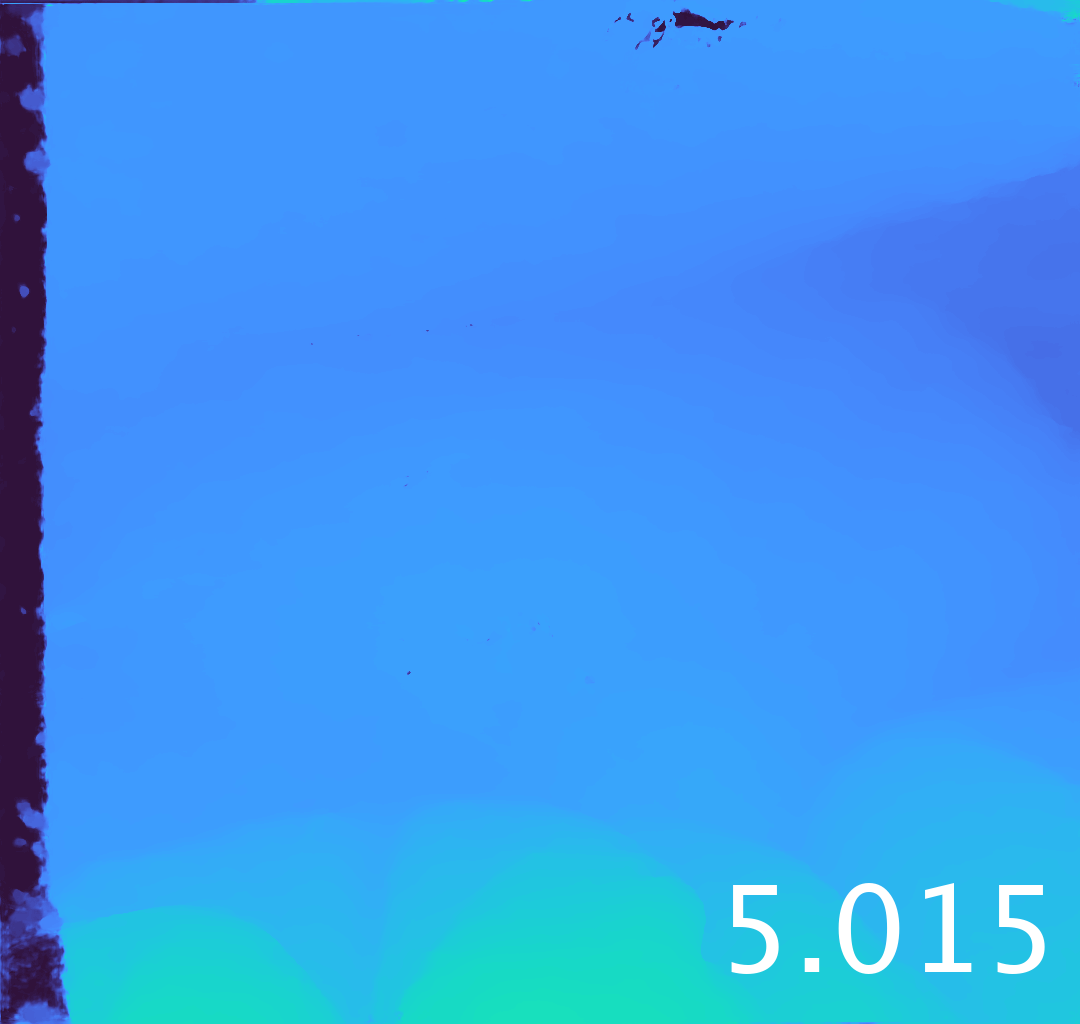} & 
    \includegraphics[width=0.195\linewidth]{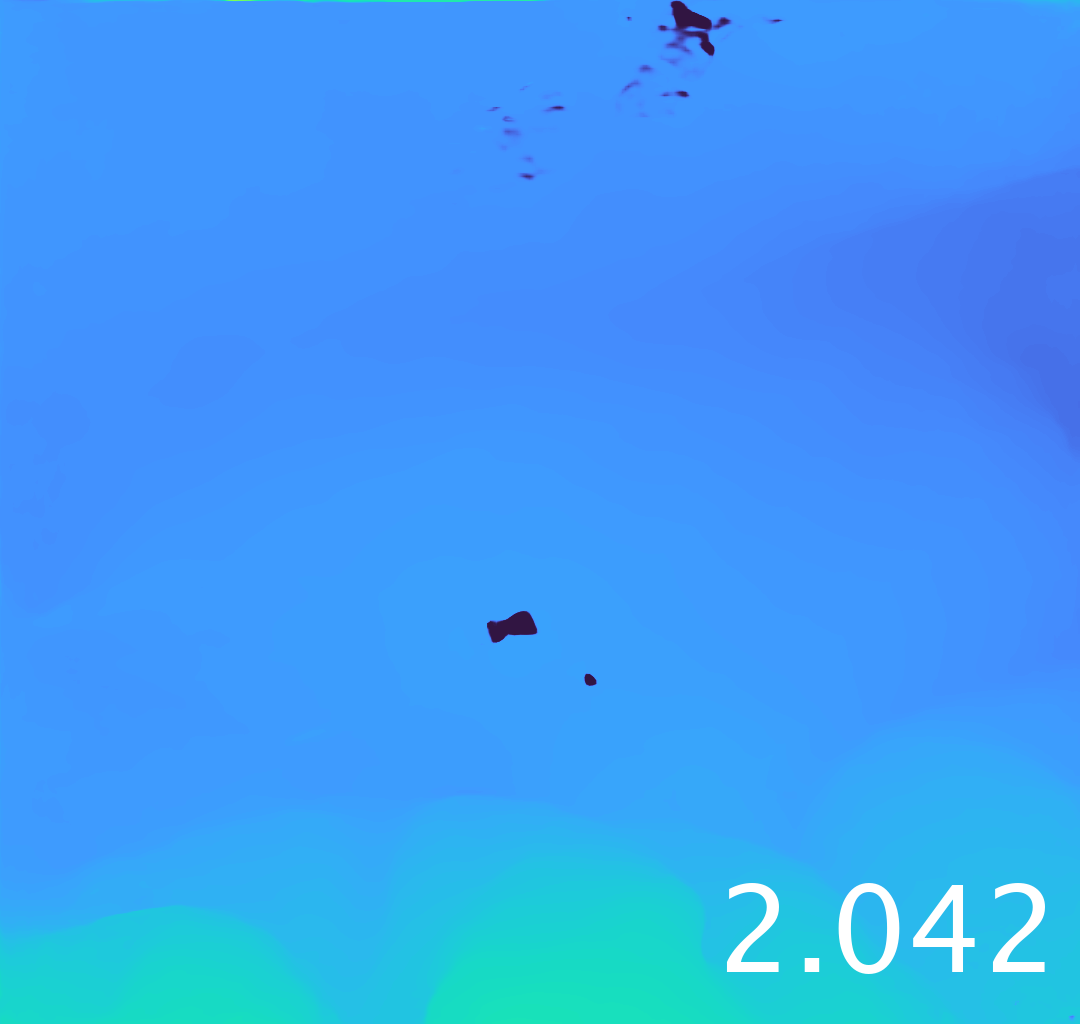} &
    \includegraphics[width=0.195\linewidth]{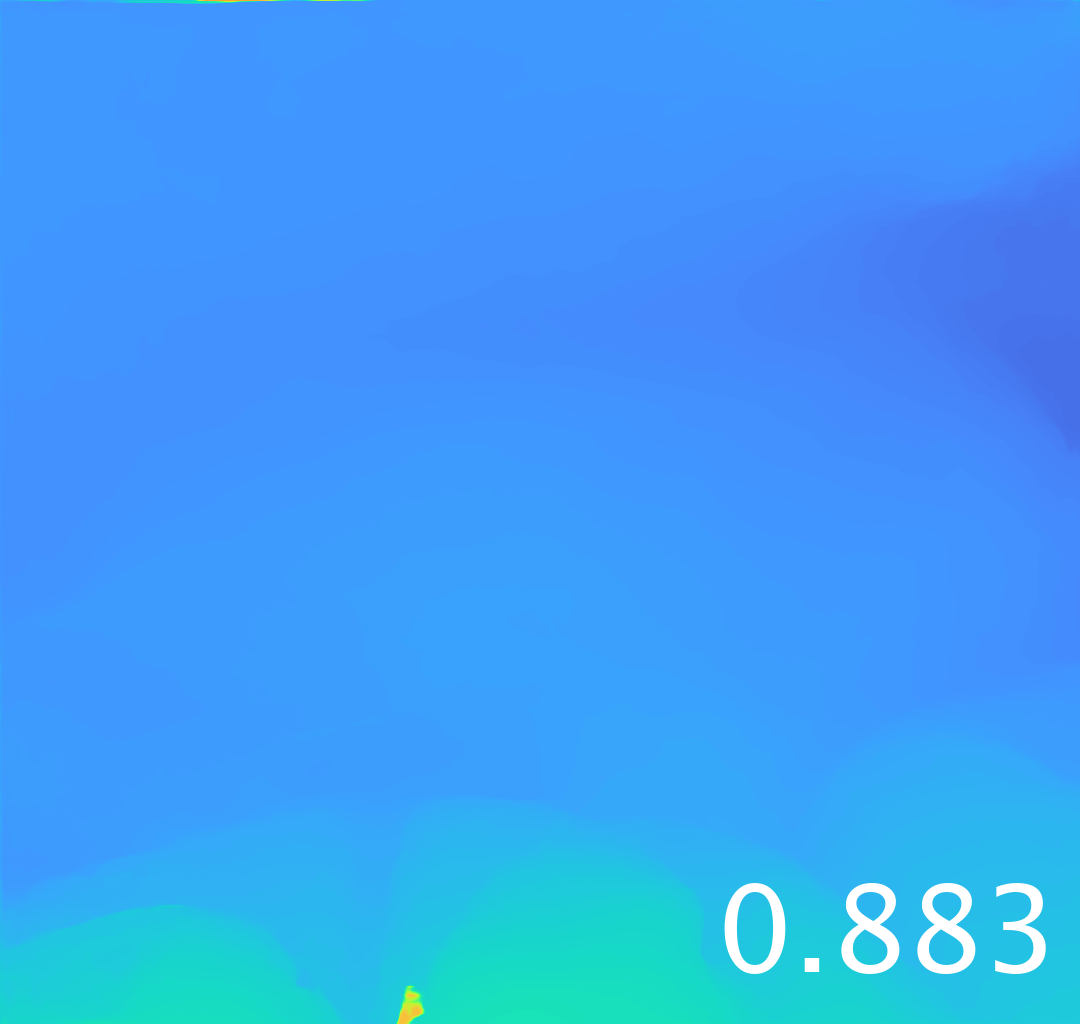} \\
    \small{Input image} & \small{GT}  &\small{STTR \cite{li2020revisiting} } & LEAStereo \cite{huang2021self} & \small{HybridStereoNet}
    \\
    \end{tabular}
    \caption{Qualitative results with bad 3.0 value on the \textit{Test-19} set. Our model predicts dense fine-grained details even for occlusion areas.}
    \label{fig:scared}
\end{figure}

\subsection{Results}
\textbf{SCARED2019.}  
We summarized the evaluation results on \textit{Test-Original} in Table \ref{tab:scared_depth}, including methods reported in the challenge summary paper \cite{allan2021stereo}. We also provided unsupervised methods from \cite{huang2021self} in the supplementary material. Note that our model never seen the training set. 
As shown in Table \ref{tab:scared_small}, our results show an improvement compared with the state-of-the-art Pure CNN method LEAStereo and a transformer-based method STTR \cite{li2020revisiting} on our reorganized \textit{Test-19} set. Please refer to supplementary material for more results on non-occluded areas.

\noindent \textbf{dVPN.} As shown in Table \ref{tab:dvpn}, our results are better than other competitors. Noting that DSSR \cite{long2021dssr} opts for the same structure with STTR~\cite{li2020revisiting}. Several unsupervised methods, \eg~SADepth \cite{huang2021self}, are not included in this table as they are trained with reconstruction losses which will lead to a high value of the evaluated SSIM metric. However, for the sake of completeness, we provide their results in supplementary material. 

\begin{table*}[!ht]
\scriptsize
\centering
\tabcolsep=0.2cm
\caption{Evaluation on the \textit{dVPN} test set (↑ means higher is better). We directly report results of ELAS and SPS from \cite{huang2021self}. Results of E-DSSR and DSSR are from \cite{long2021dssr}.} 
\begin{tabular}{l c c c c c }
\toprule
Method & Training & Mean SSIM $\uparrow$  & Mean PSNR $\uparrow$ \\
\hline
ELAS \cite{geiger2010efficient}  & No training & 47.3  & - \\
SPS \cite{yamaguchi2014efficient}  & No training & 54.7 & - \\
E-DSSR \cite{long2021dssr} & No training & 41.97$\pm$7.32 & ${13.09 \pm 2.14}$  \\ 
DSSR \cite{long2021dssr} & No training &  ${42.41 \pm 7.12}$  &  12.85 $\pm$ 2.03 \\ 
LEAStereo~\cite{cheng2020hierarchical}  & No training & 55.67 & 15.25 \\ 
HybridStereoNet  & No training & \textbf{56.98} & \textbf{15.45} \\ 
\bottomrule
\end{tabular}
\label{tab:dvpn} 
\end{table*}

\begin{figure}[t]
\centering
\tabcolsep=0.02cm
\renewcommand{\arraystretch}{1.0}
\begin{tabular}{c c c c}
\includegraphics[width=0.225\linewidth]{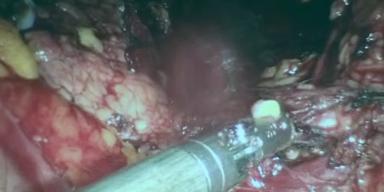} &
\includegraphics[width=0.225\linewidth]{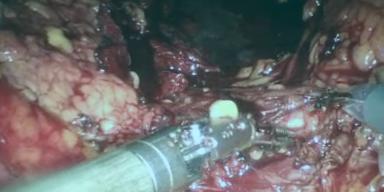} &
\includegraphics[width=0.225\linewidth]{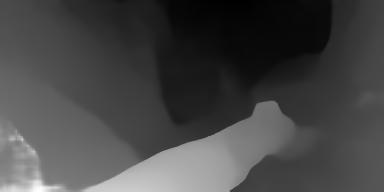} &
\includegraphics[width=0.225\linewidth]{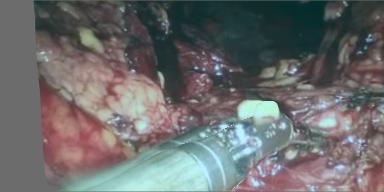}
\\
\small{Left} & \small{Right} & \small{Predict Disp}  &\small{ Reconstructed Left}
\end{tabular}
\caption{\small{{Quantitative results on the dVPN dataset}. The invalid areas on the left side of the reconstructed image are the occluded areas.}}
\label{fig:dVPN}
\end{figure}

\section{Conclusion}
\label{sec:conclusion}
In this paper, we extensively investigated the effect of transformers for laparoscopic stereo matching in terms of loss landscapes,  projected learning trajectories,  in-domain/cross-domain accuracy, and proposed a new hybrid stereo matching framework. We empirically found that for laparoscopic stereo matching, using transformers to learn feature presentations and CNNs to aggregate matching costs can lead to faster convergence, higher accuracy and better generalization. Our proposed HybridStereoNet surpasses state-of-the-art methods on SCARED2019 and dVPN datasets.

\bibliographystyle{splncs04}
\bibliography{mybibliography}

\begin{thebibliography}{10}
\providecommand{\url}[1]{\texttt{#1}}
\providecommand{\urlprefix}{URL }
\providecommand{\doi}[1]{https://doi.org/#1}

\bibitem{allan2021stereo}
Allan, M., Mcleod, J., Wang, C., Rosenthal, J.C., Hu, Z., Gard, N., Eisert, P.,
  Fu, K.X., Zeffiro, T., Xia, W., et~al.: Stereo correspondence and
  reconstruction of endoscopic data challenge. arXiv preprint arXiv:2101.01133
  (2021)

\bibitem{cartucho2021visionblender}
Cartucho, J., Tukra, S., Li, Y., S.~Elson, D., Giannarou, S.: Visionblender: a
  tool to efficiently generate computer vision datasets for robotic surgery.
  CMBBE: Imaging \& Visualization  \textbf{9}(4),  331--338 (2021)

\bibitem{chaudhari2019entropy}
Chaudhari, P., Choromanska, A., Soatto, S., LeCun, Y., Baldassi, C., Borgs, C.,
  Chayes, J., Sagun, L., Zecchina, R.: Entropy-sgd: Biasing gradient descent
  into wide valleys. Journal of Statistical Mechanics: Theory and Experiment
  \textbf{2019}(12),  124018 (2019)

\bibitem{cheng2019noise}
Cheng, X., Zhong, Y., Dai, Y., Ji, P., Li, H.: Noise-aware unsupervised deep
  lidar-stereo fusion. In: CVPR (2019)

\bibitem{cheng2020hierarchical}
Cheng, X., Zhong, Y., Harandi, M., Dai, Y., Chang, X., Li, H., Drummond, T.,
  Ge, Z.: Hierarchical neural architecture search for deep stereo matching. In:
  NeurIPS. vol.~33 (2020)

\bibitem{chong2021virtual}
Chong, N., Si, Y., Zhao, W., Zhang, Q., Yin, B., Zhao, Y.: Virtual reality
  application for laparoscope in clinical surgery based on siamese network and
  census transformation. In: MICAD. pp. 59--70. Springer (2021)

\bibitem{dosovitskiy2020image}
Dosovitskiy, A., Beyer, L., Kolesnikov, A., Weissenborn, D., Zhai, X.,
  Unterthiner, T., Dehghani, M., Minderer, M., Heigold, G., Gelly, S., et~al.:
  An image is worth 16x16 words: Transformers for image recognition at scale.
  arXiv preprint arXiv:2010.11929  (2020)

\bibitem{geiger2010efficient}
Geiger, A., Roser, M., Urtasun, R.: Efficient large-scale stereo matching. In:
  ACCV. pp. 25--38. Springer (2010)

\bibitem{hore2010image}
Hore, A., Ziou, D.: Image quality metrics: Psnr vs. ssim. In: 2010 20th
  international conference on pattern recognition. pp. 2366--2369. IEEE (2010)

\bibitem{huang2021self}
Huang, B., Zheng, J.Q., Nguyen, A., Tuch, D., Vyas, K., Giannarou, S., Elson,
  D.S.: Self-supervised generative adversarial network for depth estimation in
  laparoscopic images. In: MICCAI. pp. 227--237. Springer (2021)

\bibitem{keskar2016large}
Keskar, N.S., Mudigere, D., Nocedal, J., Smelyanskiy, M., Tang, P.T.P.: On
  large-batch training for deep learning: Generalization gap and sharp minima.
  ICLR  (2017)

\bibitem{li2018visualizing}
Li, H., Xu, Z., Taylor, G., Studer, C., Goldstein, T.: Visualizing the loss
  landscape of neural nets. NeurIPS  \textbf{31} (2018)

\bibitem{li2020revisiting}
Li, Z., Liu, X., Drenkow, N., Ding, A., Creighton, F.X., Taylor, R.H.,
  Unberath, M.: Revisiting stereo depth estimation from a sequence-to-sequence
  perspective with transformers. In: ICCV. pp. 6197--6206 (October 2021)

\bibitem{lipson2021raft}
Lipson, L., Teed, Z., Deng, J.: {RAFT-Stereo: Multilevel Recurrent Field
  Transforms for Stereo Matching}. arXiv preprint arXiv:2109.07547  (2021)

\bibitem{liu2021swin}
Liu, Z., Lin, Y., Cao, Y., Hu, H., Wei, Y., Zhang, Z., Lin, S., Guo, B.: Swin
  transformer: Hierarchical vision transformer using shifted windows. In: ICCV.
  pp. 10012--10022 (2021)

\bibitem{liu2021video}
Liu, Z., Ning, J., Cao, Y., Wei, Y., Zhang, Z., Lin, S., Hu, H.: Video swin
  transformer. arXiv preprint arXiv:2106.13230  (2021)

\bibitem{long2021dssr}
Long, Y., Li, Z., Yee, C.H., Ng, C.F., Taylor, R.H., Unberath, M., Dou, Q.:
  E-dssr: Efficient dynamic surgical scene reconstruction with
  transformer-based stereoscopic depth perception. In: MICCAI. pp. 415--425.
  Springer (2021)

\bibitem{Mayer2016CVPR}
Mayer, N., Ilg, E., Hausser, P., Fischer, P., Cremers, D., Dosovitskiy, A.,
  Brox, T.: A large dataset to train convolutional networks for disparity,
  optical flow, and scene flow estimation. In: CVPR. pp. 4040--4048 (2016)

\bibitem{Menze2015CVPR}
Menze, M., Geiger, A.: Object scene flow for autonomous vehicles. In: CVPR
  (2015)

\bibitem{nicolau2011augmented}
Nicolau, S., Soler, L., Mutter, D., Marescaux, J.: Augmented reality in
  laparoscopic surgical oncology. Surgical oncology  \textbf{20}(3),  189--201
  (2011)

\bibitem{overley2017navigation}
Overley, S.C., Cho, S.K., Mehta, A.I., Arnold, P.M.: Navigation and robotics in
  spinal surgery: where are we now? Neurosurgery  \textbf{80}(3S),  S86--S99
  (2017)

\bibitem{zhen2022cosformer}
Qin, Z., Sun, W., Deng, H., Li, D., Wei, Y., Lv, B., Yan, J., Kong, L., Zhong,
  Y.: cosformer: Rethinking softmax in attention. In: ICLR (2022)

\bibitem{scharstein2014high}
Scharstein, D., Hirschm{\"u}ller, H., Kitajima, Y., Krathwohl, G.,
  Ne{\v{s}}i{\'c}, N., Wang, X., Westling, P.: High-resolution stereo datasets
  with subpixel-accurate ground truth. In: German conference on pattern
  recognition. pp. 31--42. Springer (2014)

\bibitem{scholkopf1998nonlinear}
Sch{\"o}lkopf, B., Smola, A., M{\"u}ller, K.R.: Nonlinear component analysis as
  a kernel eigenvalue problem. Neural computation  \textbf{10}(5),  1299--1319
  (1998)

\bibitem{Sun2022}
Sun, W., Qin, Z., Deng, H., Wang, J., Zhang, Y., Zhang, K., Barnes, N.,
  Birchfield, S., Kong, L., Zhong, Y.: Vicinity vision transformer. In: arxiv.
  p. 2206.10552 (2022)

\bibitem{Wang_2021_CVPR}
Wang, J., Zhong, Y., Dai, Y., Birchfield, S., Zhang, K., Smolyanskiy, N., Li,
  H.: Deep two-view structure-from-motion revisited. In: CVPR. pp. 8953--8962
  (June 2021)

\bibitem{NEURIPS2020_add5aebf}
Wang, J., Zhong, Y., Dai, Y., Zhang, K., Ji, P., Li, H.: Displacement-invariant
  matching cost learning for accurate optical flow estimation. In: NeurIPS
  (2020)

\bibitem{wang2021pyramid}
Wang, W., Xie, E., Li, X., Fan, D.P., Song, K., Liang, D., Lu, T., Luo, P.,
  Shao, L.: Pyramid vision transformer: A versatile backbone for dense
  prediction without convolutions. In: ICCV. pp. 568--578 (2021)

\bibitem{wang2004image}
Wang, Z., Bovik, A.C., Sheikh, H.R., Simoncelli, E.P.: Image quality
  assessment: from error visibility to structural similarity. TIP
  \textbf{13}(4),  600--612 (2004)

\bibitem{yamaguchi2014efficient}
Yamaguchi, K., McAllester, D., Urtasun, R.: Efficient joint segmentation,
  occlusion labeling, stereo and flow estimation. In: ECCV. pp. 756--771.
  Springer (2014)

\bibitem{ye2017self}
Ye, M., Johns, E., Handa, A., Zhang, L., Pratt, P., Yang, G.Z.: Self-supervised
  siamese learning on stereo image pairs for depth estimation in robotic
  surgery. arXiv preprint arXiv:1705.08260  (2017)

\bibitem{zhong2017selfsupervised}
Zhong, Y., Dai, Y., Li, H.: Self-supervised learning for stereo matching with
  self-improving ability (2017)

\bibitem{zhongicpr18}
Zhong, Y., Dai, Y., Li, H.: 3d geometry-aware semantic labeling of outdoor
  street scenes. In: ICPR (2018)

\bibitem{Zhong_2018_ECCV_2}
Zhong, Y., Dai, Y., Li, H.: Stereo computation for a single mixture image. In:
  ECCV (September 2018)

\bibitem{Zhong_2019_CVPR}
Zhong, Y., Ji, P., Wang, J., Dai, Y., Li, H.: Unsupervised deep epipolar flow
  for stationary or dynamic scenes. In: CVPR (2019)

\bibitem{Zhong_2018_ECCV}
Zhong, Y., Li, H., Dai, Y.: Open-world stereo video matching with deep rnn. In:
  ECCV (2018)

\bibitem{Zhong_2022_IJCV}
Zhong, Y., Loop, C.T., Byeon, W., Birchfield, S., Dai, Y., Zhang, K., Kamenev,
  A., Breuel, T.M., Li, H., Kautz, J.: Displacement-invariant cost computation
  for stereo matching. In: IJCV (March 2022)

\end{thebibliography}

\end{document}